\documentclass{article}

\usepackage{arxiv}

\usepackage[utf8]{inputenc} 
\usepackage[T1]{fontenc}    
\usepackage{hyperref}       
\usepackage{url}            
\usepackage{booktabs}       
\usepackage{amsfonts}       
\usepackage{nicefrac}       
\usepackage{microtype}      
\usepackage{amsmath}
\usepackage{cleveref}       
\usepackage{graphicx}
\usepackage{natbib}
\usepackage{doi}

\title{Symbolic Synthesis of Neural Networks}

\date{}

\author{Eli Whitehouse \\
	New York, NY 10024 \\
	\texttt{eliw55@gmail.com} \\
}

\renewcommand{\headeright}{}
\renewcommand{\undertitle}{}
\renewcommand{\shorttitle}{Symbolic Synthesis of Neural Networks}

\hypersetup{
pdftitle={Symbolic Synthesis of Neural Networks},
pdfauthor={Eli Whitehouse},
pdfkeywords={neural networks, symbolic programs, graph neural networks, library learning, distributional program search},
}

\begin{document}
\maketitle

\begin{abstract}
Neural networks adapt very well to distributed and continuous representations, but struggle to generalize from small amounts of data.  Symbolic systems commonly achieve data efficient generalization by exploiting modularity to benefit from local and discrete features of a representation.  These features allow symbolic programs to be improved one module at a time and to experience combinatorial growth in the values they can successfully process.  However, it is difficult to design a component that can be used to form symbolic abstractions and which is adequately overparametrized to learn arbitrary high-dimensional transformations.  I present Graph-based Symbolically Synthesized Neural Networks (G-SSNNs), a class of neural modules that operate on representations modified with synthesized symbolic programs to include a fixed set of local and discrete features.  I demonstrate that the choice of injected features within a G-SSNN module modulates the data efficiency and generalization of baseline neural models, creating predictable patterns of both heightened and curtailed generalization.  By training G-SSNNs, we also derive information about desirable semantics of symbolic programs without manual engineering.  This information is compact and amenable to abstraction, but can also be flexibly recontextualized for other high-dimensional settings.  In future work, I will investigate data efficient generalization and the transferability of learned symbolic representations in more complex G-SSNN designs based on more complex classes of symbolic programs.  Experimental code and data are available \href{https://github.com/shlomenu/symbolically_synthesized_networks}{here}.
\end{abstract}


\section{Introduction}
\label{sec:intro}

Most conventional modes of human communication naturally occur in a high-dimensional medium such as text, audio, or images.  When processing these media, slight errors can easily accrue across many dimensions of the input where features are distributed.  With adequate data, neural networks adapt effectively to these patterns by adjusting many interdependent real-valued parameters.  In their basic form, however, neural models are data inefficient.  In settings where more data cannot be sourced, pretraining has arisen as the most general and data-driven answer to this challenge.  Pretraining allows practitioners to repurpose data that is not specifically suited to a task to enrich a representation or initialize parameters of a neural model.  While pretraining grants a model exposure to out-of-distribution data which may be irrelevant or inappropriate to the task at hand, the benefits of this exposure can outweigh the costs.

In symbolic systems, modularity can allow for greater data efficiency and generalization in the presence of local and discrete features.  By examining the particular dimensions of an input which cause systems to fail, developers can trace issues back to specific modules.  When a failure in one module is corrected, the system benefits from this across all combinations of values and dimensions, witnessing potentially exponential growth in the number of inputs on which it succeeds.  In contrast to pretraining, the use of modular functionalities to process local and discrete features offers a direct solution to the challenge of data efficient generalization.  However, it is difficult to design a unit that is appropriate both for learning arbitrary high-dimensional transformations, and for composing symbolic abstractions, as the overparametrization that gives neural networks their flexibility also makes their semantics unstable.  This makes it difficult to judge which combinations of primitive operations are useful enough to merit the creation of a new abstractions.

I present a novel approach based on Graph-based Symbolically Synthesized Neural Networks (G-SSNNs), a form of neural module in which symbolic abstraction and gradient-based learning play complementary roles.  In a G-SSNN, the output of a symbolic program is used to determine a set of local and discrete features that are injected into the representations processed by the network.  The symbolic program is chosen by an evolutionary algorithm, which is used to develop and evaluate a population of G-SSNNs with varying designs.  Through the use of distributional program search and library learning---a mechanism for identifying useful new abstractions---the evolutionary algorithm may influence both the content of injected features and the degree of locality they exhibit.  In this work, I apply evolving populations of G-SSNNs to a binary prediction task over high-dimensional data that humans find simple and easy to interpret from handfuls of examples.  I show that these populations exhibit a reliable pattern of both heightened and curtailed generalization relative to baseline models after training on small quantities of data.

\section{G-SSNNs}

In a G-SSNN, the output of a symbolic program fixes a mechanism that injects a set of local and discrete features into the input.  A graph may be used to represent not just these individual features as node and edge representations, but also their relationships to each other.  After a symbolic program has generated a graph $g \in G_k$ with symbolic node and edge representations $v \in \mathbb{N}^k$, we may embed arbitrary input vectors into this structure piecewise with a function $e : \mathbb{N}^k \times \mathbb{R}^m \to \mathbb{R}^q$.  Each symbolic feature thus provides an instruction as to how a region of the input vector should be transformed.  With $\texttt{graph\_map}_e : G_k \times \mathbb{R}^m \to G^\prime_q$, a function that applies $e$ for each symbolic feature $v \in g$, we compute a graph $g^\prime \in G^\prime_q$ with transformed, real-valued features which may be processed by a standard GNN $m_{gnn} : G^\prime_q \to \mathbb{R}^n$.  The G-SSNN $m : \mathbb{R}^m \to \mathbb{R}^n$ is then a triplet of $g$, $e$, and $m_{gnn}$ used to compute 
\begin{equation*}
    m(x) = m_{gnn}(\texttt{graph\_map}_e(g, x)).
\end{equation*}  
Provided that $e$ is differentiable with respect to $x$, the G-SSNN $m$ may be augmented with a readout or global pooling function to create a generic neural module.  There are many definitions of $G_k$ and $e$ practitioners may adopt depending on the intended role of the injected features.  The definitions used in this work are presented in section \ref{sec:g-ssnn-design} of the appendix.

\section{Evolutionary Framework}

While the GNN of a G-SSNN can be trained with standard gradient-based optimization techniques, its graph structure is a hyperparameter whose values must be explored with search.  To conduct this search, I repurpose the cycle of library learning and distributional program search pioneered by the DreamCoder system \citep{dreamcoder-bootstrapping-inductive-program-synthesis-with-wake-sleep-library-learning} within an evolutionary framework, incorporating the improved STITCH library learning tool of \citet{stitch} and heap search algorithm of \citet{heap-search}.  As in DreamCoder, I perform rounds of library learning and distributional program search, after which I evaluate the current crop of programs on the task.  Unlike symbolic programs, however, G-SSNNs may exhibit widely-varying degrees of generalization.  As such, it is not appropriate to take their performance on a handful of training examples as indicative of goodness of fit.  Rather, a heuristic must be applied to select those symbolic programs whose corresponding G-SSNNs are expected to generalize best based on their training performance.  G-SSNNs, along with such a heuristic, may be understood as generalizing the evaluation mechanisms used for symbolic programs within the DreamCoder system.  In this work, I develop a single population of G-SSNNs for application to a single task.  However, in principle, nothing prevents the multitask Bayesian program learning framework of the DreamCoder system from being applied to the development of multiple populations of G-SSNNs.

After each evolutionary step, a new population is produced for each task.  At the next step of library learning, novel primitives are judged according to which provide the greatest degree of compression of this corpus (i.e. reduction in the number of primitives needed to express a program).  The creation of new primitives is therefore guided by the pruning of the population that has occurred at previous stages.  By inferring the distribution of unigrams of primitives in the current population, we may guide search towards programs that utilize these primitives at similar frequencies.  Through these mechanisms, the choice of heuristic influences the evolution of future G-SSNN designs.  Though no heuristic is best in all cases, the class of heuristics that discard models with the Lowest Training Performance (LTP) is simple and agreeable for a wide variety of applications.  In this work, I utilize a heuristic I refer to as rank-LTP-50, which directs that the bottom half of G-SSNNs in terms of ranked training performance be discarded at each step of evolution.  For additional details on the search space explored with these techniques, see section \ref{sec:symbolic-program-design} of the appendix.

\section{Related Work}

Neural Architecture Search (NAS) is the process of using automated methods to explore the discrete space of neural network topologies.  In a G-SSNN $m$, the symbolic program may be considered to influence the topology of the model through its role in structuring the applications of graph convolutions within $m_{gnn}$, and also through its role in distinguishing dimensions of the input through the embedding function $e$.  However, despite variation in the design of their search space \citep{alpha-nas-property-guided}, search strategies \citep{automl-zero}, and model evaluation strategies \citep{zero-cost-nas}, all approaches I reviewed assume that parameters will be learned with gradient-based optimization before each model is assessed.  This differs from G-SSNNs, as the embedding function described in section \ref{sec:g-ssnn-design} includes a term whose function is equivalent to that of a bias parameter, yet remains static throughout gradient-based learning.  The act of adding features to a representation through a static mechanism seems more analogous to an act of preprocessing, as might be performed with word embeddings \citep{mikolov-2013-word-embeddings} or positional encoding \citep{attn-is-all-u-need}.  However, I found no works in which similar acts of preprocessing were designed with the help of distributional program search and library learning.  G-SSNNs draw significant inspiration from the DreamCoder system; however, I did not find examples of neurosymbolic systems in which adjacent parametric and nonparametric mechanisms are separately optimized through gradient-based learning and library learning.

\section{Experiments}

\subsection{Setup \& Hyperparameters}

To facilitate comparison, I utilize a base model architecture consisting of a transformer with a convolutional stem \citep{early-convolutions-help-transformers-see-better}.  I use subpixel convolution to perform downsampling from $128 \times 128$ to $8 \times 8$ \citep{subpixel-conv}, followed by a pointwise convolution and three residual convolutional units.  The patches are then flattened and fed to a simplified Vision Transformer (ViT) \citep{simple-vit} with 2 transformer blocks.  Each block has an MLP dimension of 512 and 4 128-dimensional attention heads.  Global average pooling is then applied to produce an output of the target dimension for baseline models, or with the dimensionality of a node/edge representation for experimental models in which the output is fed to a G-SSNN unit.

To construct G-SSNNs, I use the Graph Isomorphism Network with Edge features (GINE) of \citet{gine} as implemented in the DGL library \citep{dgl}.  I use 512-dimensional node and edge representations, which are passed through three GINE layers parametrized by an MLP with identical input and output dimensionalities.  Between GINE layers, I apply 30\% droput \citep{dropout}.  To produce the final output, I apply average pooling followed by another MLP which projects its input to the target output.

For both baseline and experimental models I use a batch size of 8, train each model for 16 epochs with the Adam optimizer \citep{adam-optimizer}, and reduce the learning rate by a factor of 1/2 when a new low in the loss has not been experienced in the first 50 batches, or in the most recent 65 batches since the learning rate was last reduced.

Across iterations of evolutionary selection, I retain a population of at most 50 and apply the rank-LTP-50 heuristic if the size of the population is greater than or equal to 25.  I run distributional program search for a 15 second period.  In the course of this run, I retain only the most likely $50 - \texttt{population\_size}$ novel programs; however, if programs are discovered that are shorter than, but semantically equivalent to, programs currently in the population, these shorter reformulations are kept and their equivalents are set aside.  I do not consider programs that consist of more than 150 primitives from the most updated DSL.  Prior to program search, I reweight the parameters of the program distribution such that the log likelihoods of the most and least likely primitives differ by no more than 0.5.  I implemented this transformation such that it preserves the relative distance of each primitive's log likelihood from the mean and therefore does not alter the DSL's total probability mass.\footnote{This and other domain-general aspects of program search are implemented in the \href{https://github.com/shlomenu/antireduce}{\texttt{antireduce}} library.}  During library learning, I perform up to three rounds of compression, each producing a primitive of arity 2 or less.  All other parameters of the STITCH compression routine were allowed to take their default values.\footnote{Explanation of these settings may be found in the \href{https://stitch-bindings.readthedocs.io/en/stable/index.html}{documentation} of the \texttt{stitch\_core} Python package.}

\begin{figure}
\centering
\begin{minipage}{.3\textwidth}
    \centering
    \includegraphics[width=5.5cm]{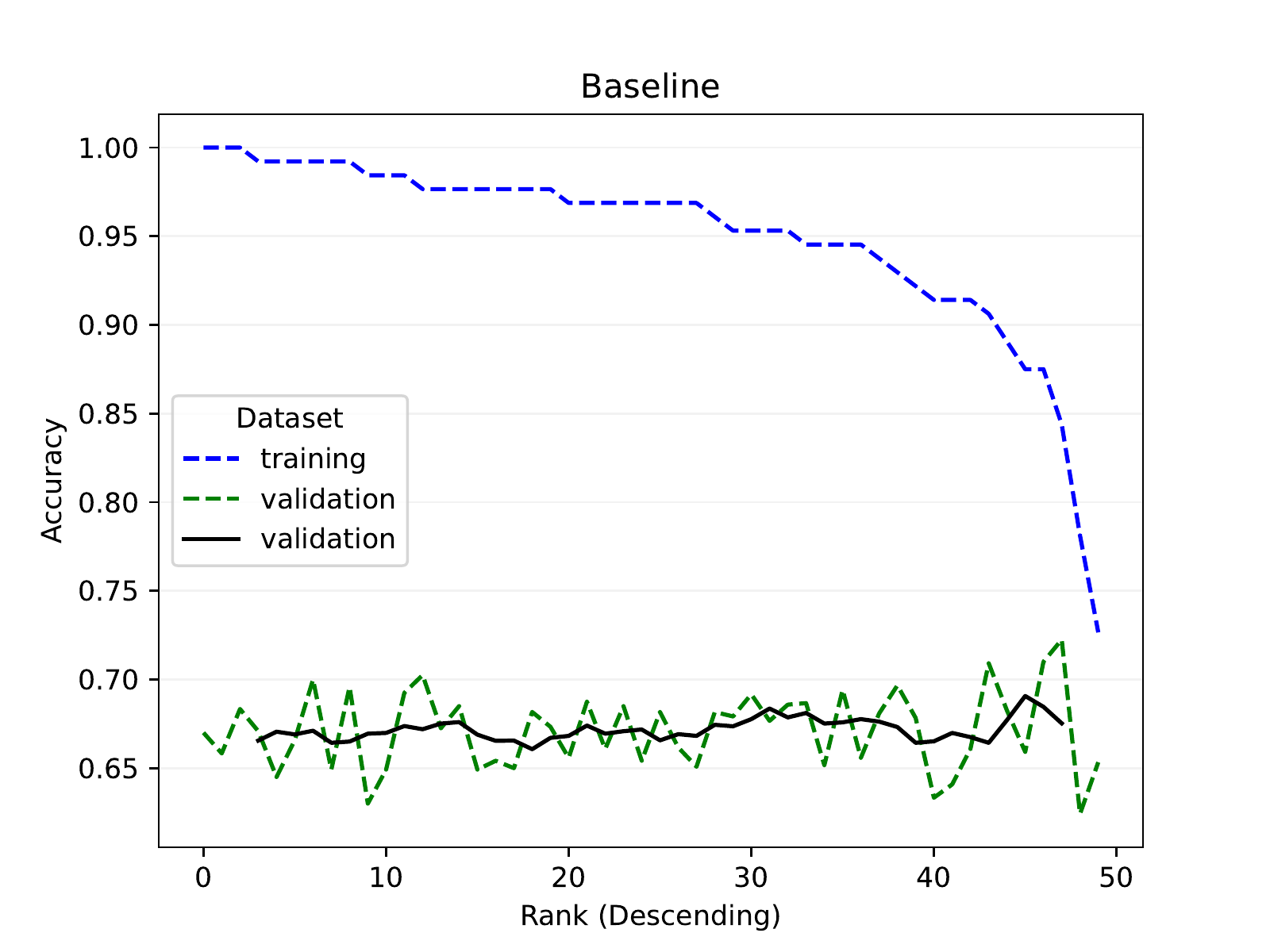}
\end{minipage}%
\begin{minipage}{.3\textwidth}
    \centering
    \includegraphics[width=5.5cm]{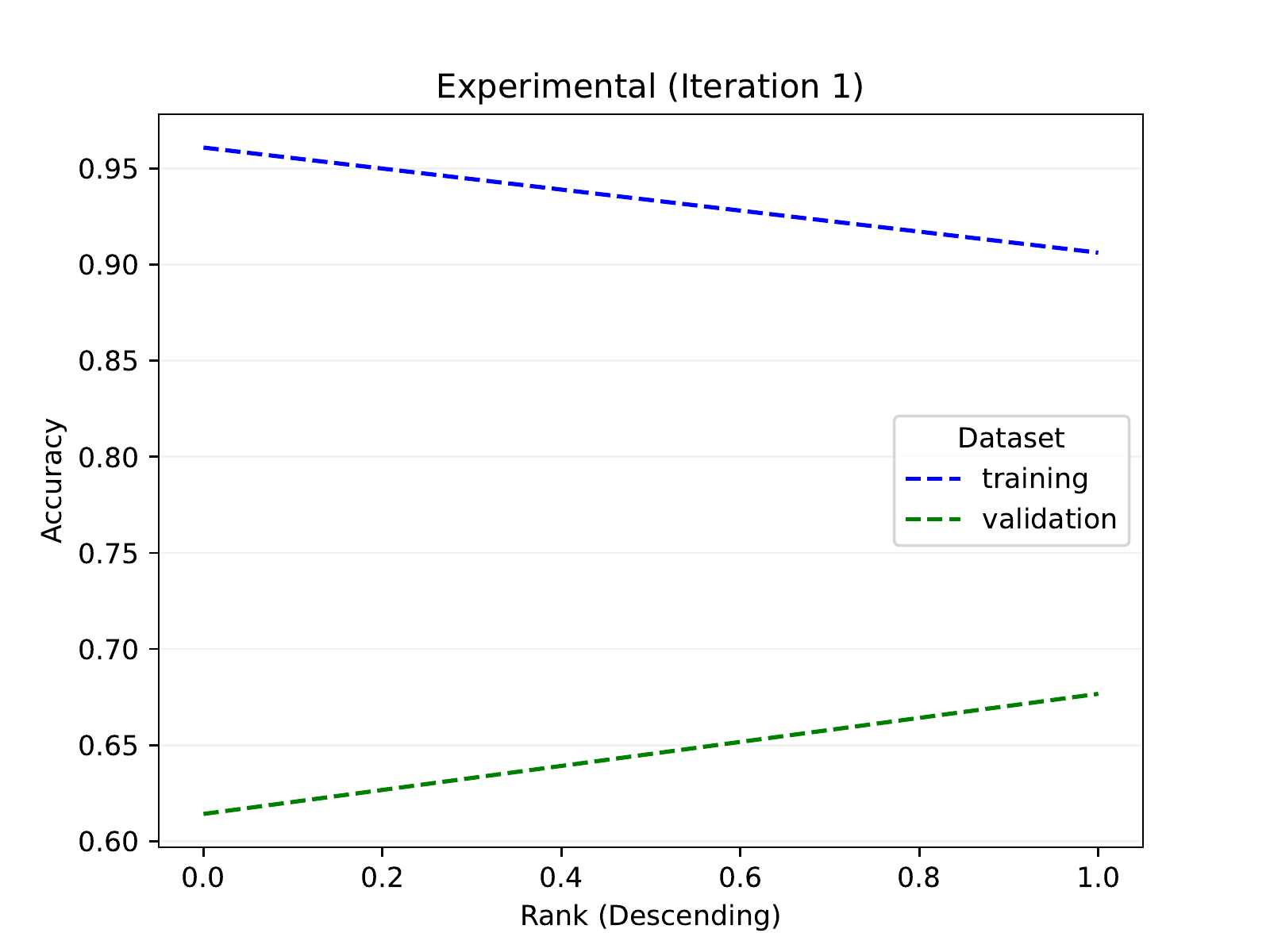}
\end{minipage}%
\begin{minipage}{.3\textwidth}
    \centering
    \includegraphics[width=5.5cm]{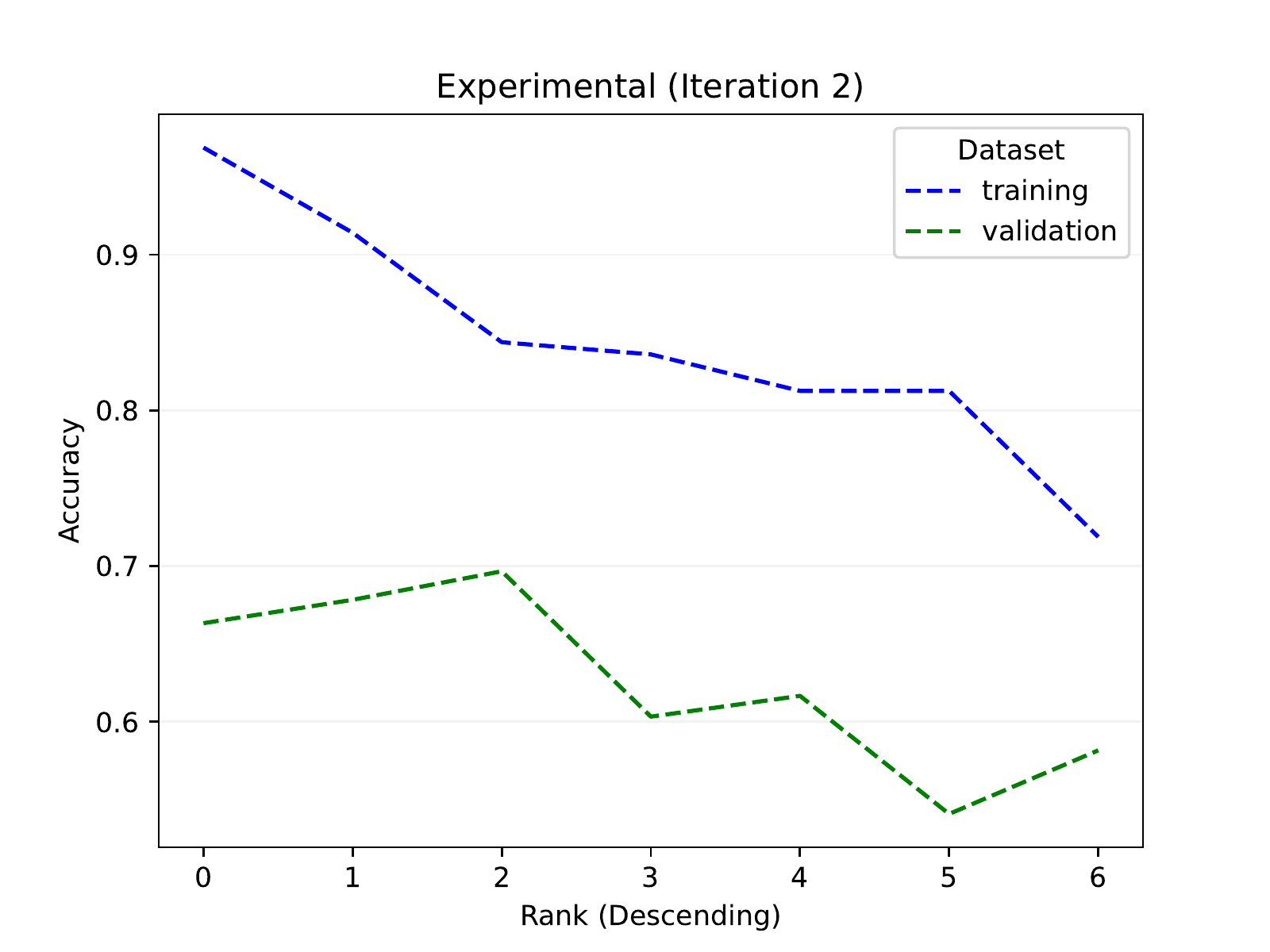}
\end{minipage}
\begin{minipage}{.3\textwidth}
    \centering
    \includegraphics[width=5.5cm]{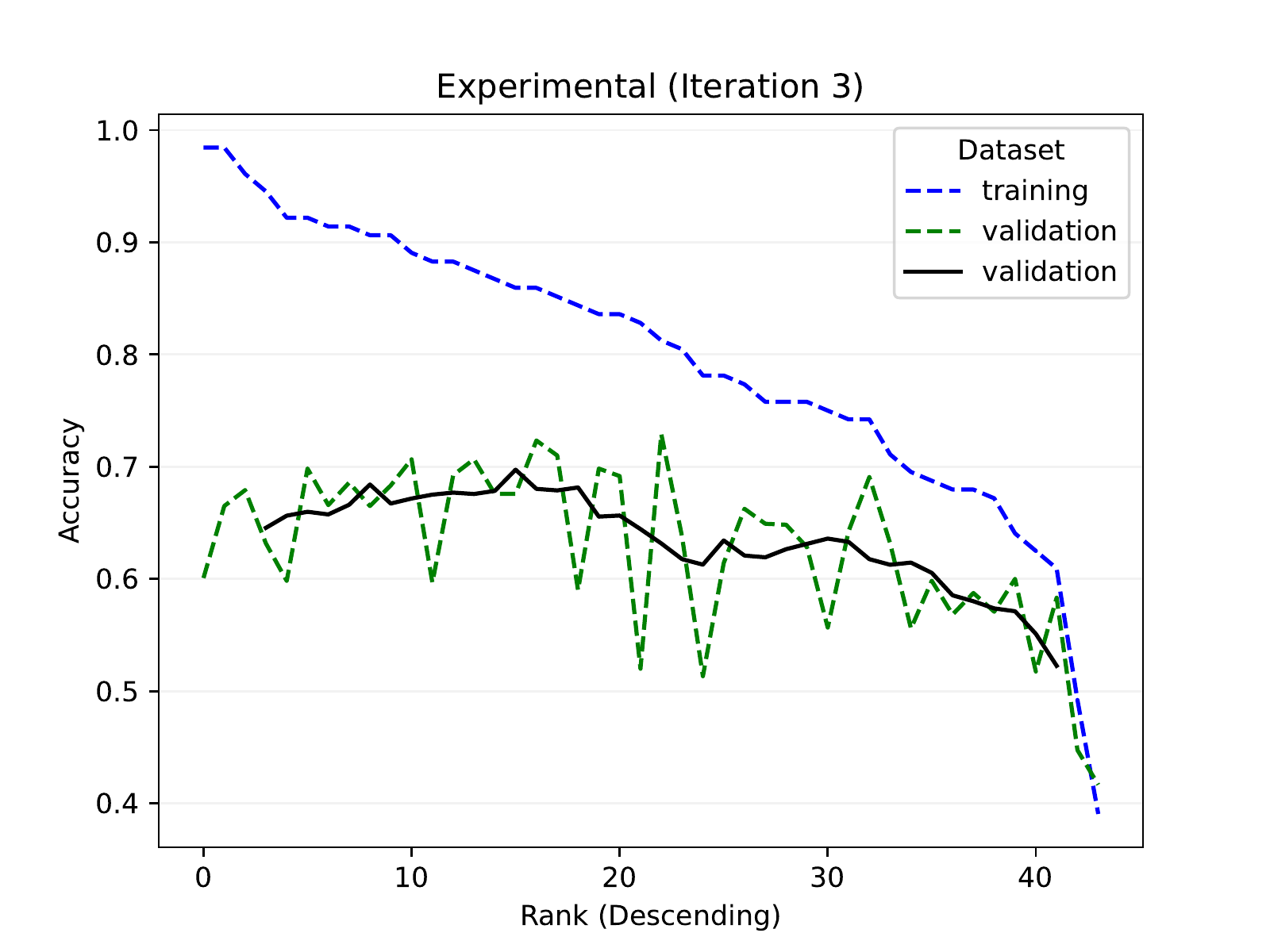}
\end{minipage}%
\begin{minipage}{.3\textwidth}
    \centering
    \includegraphics[width=5.5cm]{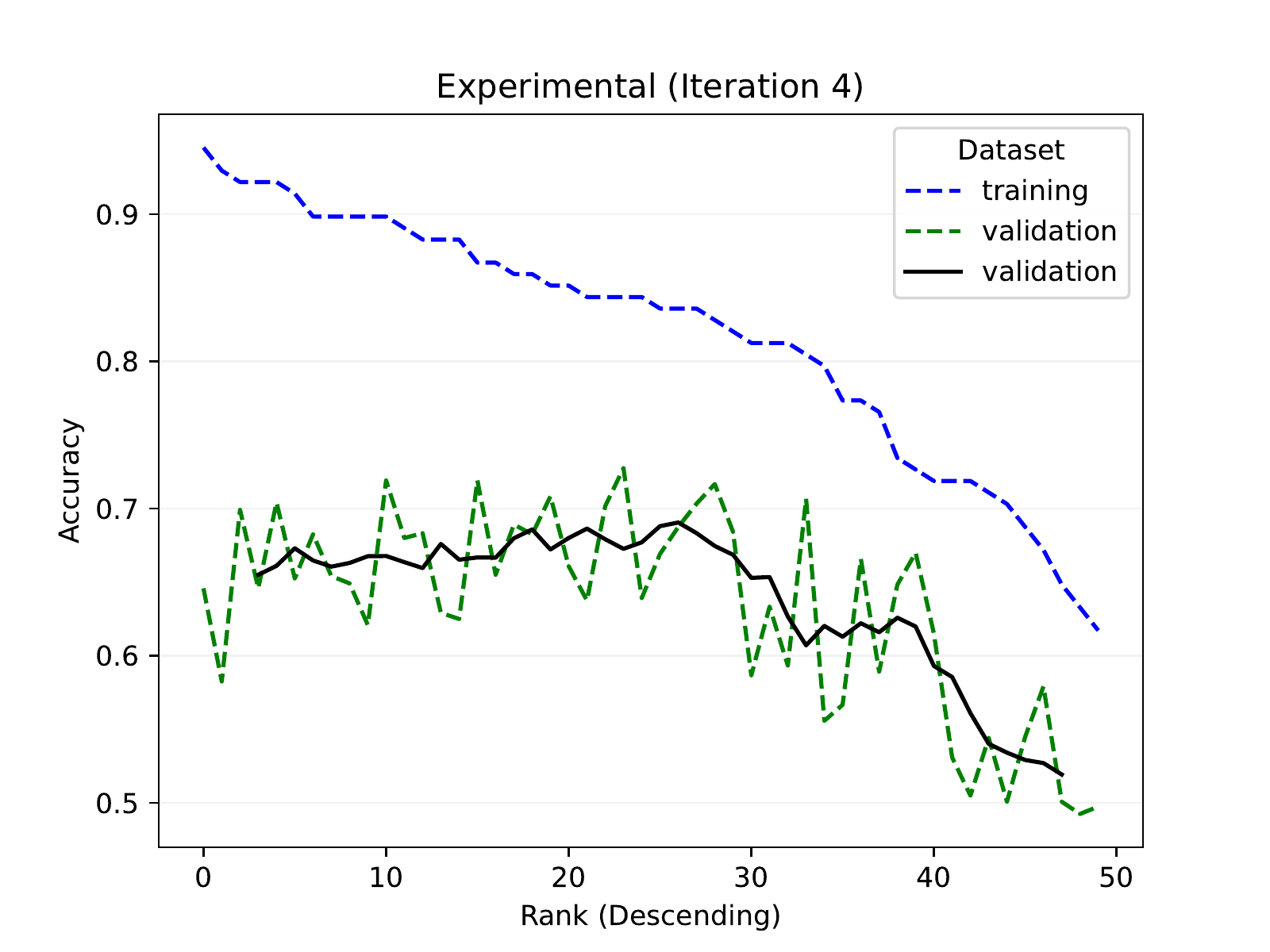}
\end{minipage}%
\begin{minipage}{.3\textwidth}
    \centering
    \includegraphics[width=5.5cm]{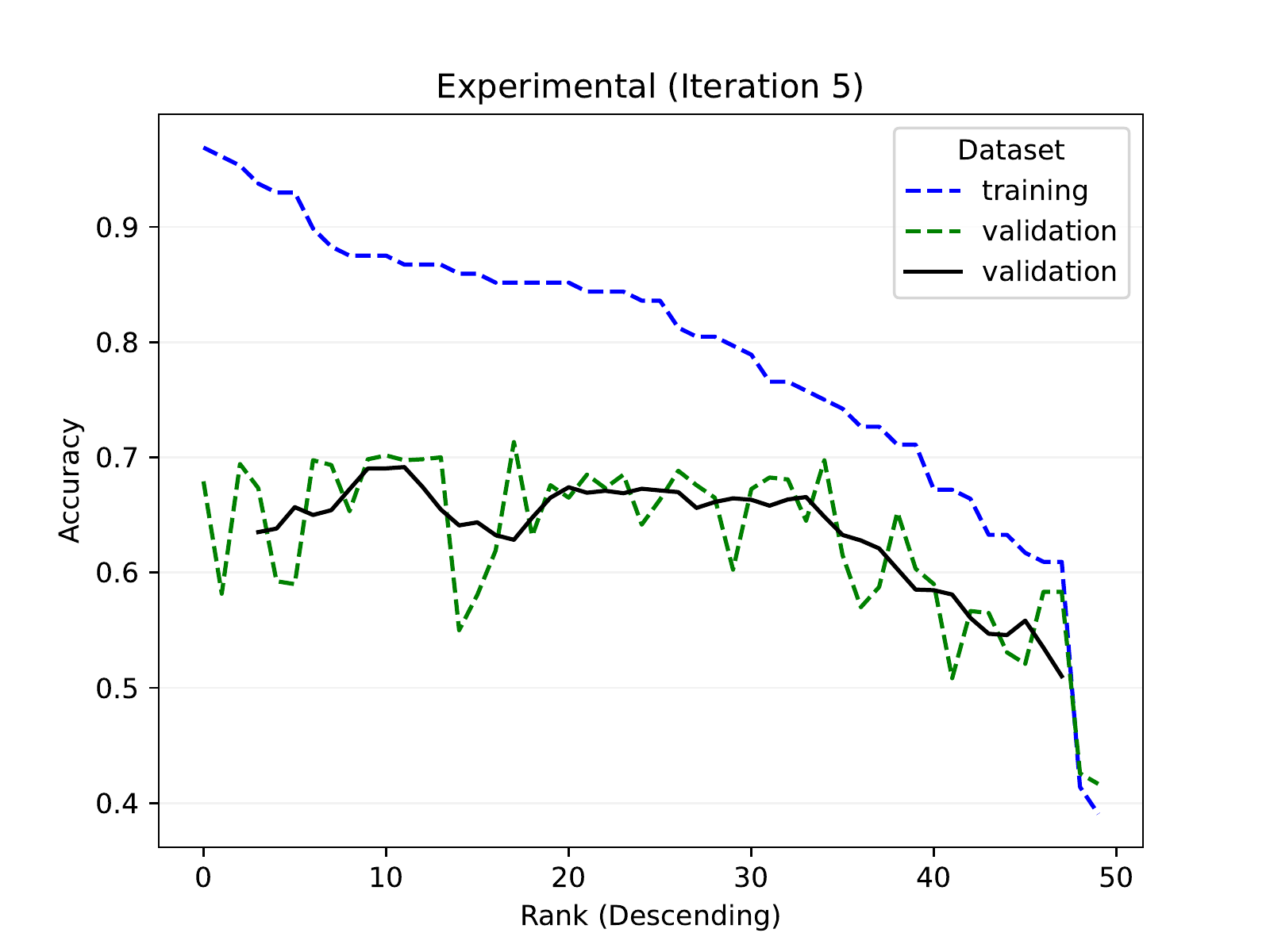}
\end{minipage}
\caption{Training and validation set performances of populations of neural networks.  Each model of a population is represented in both unsmoothed trend lines of a plot at parallel points with respect to the x-axis.  Parallel points are sorted in descending order of training set accuracy.  The baseline plot depicts the spread of training and validation set accuracies for 50 instantiations of a baseline model.  The experimental plots depict the spread of training set and validation set accuracies for populations of G-SSNNs at various stages of evolution.  At stage $n$, the population has undergone $n$ rounds of heuristic selection and program synthesis.  At iterations 1-5, the populations were of sizes 2, 7, 44, 50 and 50, respectively.  Dotted lines represent unsmoothed data and solid lines represent smoothed data with a window size of 6.  I present validation set accuracies in both smooth and unsmoothed form except where the population of models seem too small to meaningfully benefit from smoothing ($n < 15$).}
\label{fig:perf}
\end{figure}

\subsection{Dataset}

The RAVEN dataset \citep{raven-a-dataset-for-analogical-and-visual-reasoning} is an artificial dataset consisting of visual puzzles known as Raven's Progressive Matrices (RPMs).  Originated as a psychometric test, RPMs have since been extensively studied as tests of Abstract Visual Reasoning (AVR) in both symbolic and neural AI systems.  An RPM is a $9 \times 9$ matrix of panels depicting simple arrangements of polygons.  Within each row, panels exhibit one of a handful of consistent patterns with respect to each attribute--size, shape, color, angle, position, and multiplicity--of their polygons.  These patterns are easily observed by humans from relatively few examples.  In comparison, neural models typically require orders of magnitude more data to match or surpass human performance. 

Classically, the final panel of the last sequence (bottom right) is omitted and some number of candidate panels are provided, only one of which contains polygons which obey the same patterns exhibited in the first two rows for all attributes.  The goal of the solver is to distinguish the correct panel from the pool of candidates. \citet{evaluating-understanding-on-conceptual-abstraction-benchmarks} have demonstrated that several recent high-performing neural models from the RAVEN literature perform poorly when more exhaustively evaluated on their understanding of the various instantiations of an individual pattern.  This inspired the Prediction of Constant Color Pattern (PCCP) task, a binary prediction task in which a model examines a complete and correct RPM and predicts whether it exhibits constancy of polygon color across rows.  The two classes in this task--constant and non-constant--were relatively evenly-distributed among generated data without additional constraints on pattern instantiation or attribute values.\footnote{Some constraints on pattern instantiation and attribute values were necessary to ensure legibility at the desired resolution of $128 \times 128$.  Specifically, polygon size was kept large and not allowed to vary and oblique angles of polygon rotation were disallowed.  Neither of these constraints contribute to determining the color constancy of an RPM.}  Furthermore, color is unrelated to and unaffected by the settings of other attributes, so indiscernible spurious correlation with the presence or absence of other pattern-attribute pairings is unlikely, even with small data.

The RAVEN-like data used in my experiments were generated with a \href{https://pypi.org/project/raven-gen/}{reimplementation} of the original \href{https://github.com/WellyZhang/RAVEN}{RAVEN data generation code} extended to support creation of RPMs at lower resolutions than were originally accommodated.  I use a training dataset size of 128 RPMs and a validation set of 1200 RPMs.  These sizes make it easy to fit to the training data, but difficult to generalize without appropriate inductive biases.

\subsection{Results}

Figure \ref{fig:perf} presents comparative results for populations of baseline and experimental models.  Generally, experimental models exhibit a much smaller gap in performance between seen and unseen data than baseline models do.  For both baseline and experimental models, validation set accuracy is highest for models whose training set accuracy is in the range of 80-90\%.  Among baseline models less than 20\% fall in this range.  After the first iteration, 30\% or more of the population of experimental models fall in this range.  For both baseline and experimental models, validation set accuracy is lowest for models whose training set accuracy falls below 80\%.  While only one or two baseline models fall in this range, by the third iteration 20\% or more of experimental models fall in this range.  The experimental condition thus makes both peak validation set accuracies and degraded validation set accuracies more numerous, providing evidence that the use of G-SSNN modules modulates the data efficiency and generalization of the baseline model.  It also appears that among experimental models, there is a clearer relationship between training set accuracy and validation set accuracy, making it easier to select models that exhibit goodness of fit on both seen and unseen data.    

The distributions of validation set accuracies illustrated in these data suggest that it is actually models with a training set performance of a more intermediate rank that are most likely to exhibit peak generalization.  In these models, injected features raise validation set performance without lowering training set performance to a harmful degree.  In future work and applications of G-SSNNs, it may therefore be beneficial to explore the space of Intermediate Training Performance (ITP) heuristics for selecting from the population.    

\section{Discussion}

Across high-dimensional media, there are tasks that require data efficient generalization.  In these settings, modular systems may be better at responding to local and discrete features that simplify and accelerate learning.  In G-SSNNs, we stimulate modularity by injecting a fixed set of local and discrete features which the model must contextualize for the task at hand.  This allows G-SSNN modules to be trained with standard techniques.  By training G-SSNNs, we also derive information about desirable semantics of symbolic programs without manual engineering.  Unlike the knowledge contained in neural network weights, this information can be abstracted upon just as the functions of handwritten software libraries are today.  However, unlike the knowledge contained in purely symbolic systems, it can also be flexibly adapted to high-dimensional settings in ways that cannot be concisely expressed with symbols.  In advancing this paradigm, I believe that G-SSNNs capture much of what is desirable about the promise of common sense: the ability to quickly adapt and elaborate on compact knowledge in a range of perceptually-grounded circumstances.  

In future work, I will investigate data efficient generalization and the transferability of learned symbolic representations in more complex G-SSNN designs based on more complex classes of symbolic programs.  Such programs could include operations that create more varied node and edge features and access richer external sources of information to inform their values.  Based on these features, embedding functions could utilize more complex topological structures, or incorporate transformations from pretrained models.

\section{Conclusion}

I have presented G-SSNNs, a class of neural modules whose representations are modified with synthesized symbolic programs to include a fixed set of local and discrete features.  I presented the results of applying neural models containing G-SSNN modules to a binary prediction task over Raven's Progressive Matrices (RPMs), psychometric tests of Abstract Visual Reasoning (AVR) on which humans exhibit data efficient generalization.  These results demonstrate that the injected features of G-SSNNs modulate the data efficiency and performance of baseline neural models.  In addition to their potential to improve data efficiency and generalization on tasks involving high-dimensional transformations, G-SSNNs also allow us to derive information about the desirable semantics of symbolic programs without manual engineering.  This information is compact and amenable to abstraction, and can also be flexibly recontextualized for other high-dimensional settings in response to data.  In future work, I will investigate data efficient generalization and the transferability of learned symbolic representations in more complex G-SSNN designs based on more complex classes of symbolic programs.     

\bibliographystyle{unsrtnat}
\bibliography{references}  






\renewcommand\thesection{A}

\section*{Appendix}

\subsection{G-SSNN Design}
\label{sec:g-ssnn-design}

For the experiments presented above, I utilize the class $G_2$ consisting of single-component undirected graphs with node and edge features.  The symbolic features $v \in g$ for $g \in G_2$ correspond to a unique ID $v_1$ and a neighborhood rank $v_2$.  The ID of a graph element $v$ (node or edge) reflects the order in which the symbolic program created the elements of the graph.  It is given by the number of elements present in the graph at the time $v$ was created, and is therefore 0-indexed.  I denote the number of unique IDs in the graph by $d^\ast = 1 + \max_{v \in g} v_1$.  Neighborhood ranks are natural numbers in a fixed range which distinguish the edges of a particular neighborhood.  The neighborhood rank of a node is 0, while the neighborhood rank of an edge ranges from 1 to a maximum degree.  Since neighborhood ranks distinguish edges even if their endpoints are the same, the classes $G_k$ and $G^\prime_n$ may be understood as permitting multiedges.  Self-edges, however, are not allowed.  I denote the number of neighborhood ranks by $r^\ast = 1 + \max_{v \in s} v_2$.  

I utilize the embedding function 
\begin{equation*}
    e(v, x) = \texttt{tile\_expand}(x_{\texttt{idx}(v_1):\texttt{idx}(v_1 + 1)}) + \texttt{2d\_pos}(v_1, v_2)
\end{equation*}
where $\texttt{idx}(w) = \boldsymbol{1}[w \neq 0] \cdot (m \bmod l) + w \cdot l$ for $l = \lfloor m / d^\ast \rfloor$.  The function $\texttt{tile\_expand} : \mathbb{R}^t \to \mathbb{R}^q$ for variable $t$ implements the differentiable operation of concatenating $\lceil q / t \rceil$ copies of its input and truncating the result to a dimensionality of $q$. The function $\texttt{2d\_pos} : \mathbb{N}^2 \to \mathbb{R}^q$ produces computes a vector $b = \texttt{2d\_pos}(d, r)$ such that 
\begin{equation*}
    b_i = \begin{cases}
     \sin(\texttt{reorder\_1}(d, r) / 10000^{i \bmod (n / 4)}) & \hspace{5.85mm} 1 < i \leq n / 4 \\
     \cos(\texttt{reorder\_1}(d, r) / 10000^{i \bmod (n / 4)}) & \hspace{1.85mm} n / 4 < i \leq 2n / 4 \\
     \sin(\texttt{reorder\_2}(d, r) / 10000^{i \bmod (n / 4)}) & 2n / 4 < i \leq 3n / 4 \\
     \cos(\texttt{reorder\_2}(d, r) / 10000^{i \bmod (n / 4)}) & 3n / 4 < i \leq n
    \end{cases}.
\end{equation*}
The functions \texttt{reorder\_1} and \texttt{reorder\_2} are an implementation detail due to mistaken indexing which did not affect the validity of the experiment.  These functions are defined as $\texttt{reorder\_1}(d, r) = (d \cdot r^\ast + r) \bmod d^\ast$ and $\texttt{reorder\_2}(d, r) = \lfloor (d \cdot r^\ast + r) / d^\ast \rfloor$.  These functions may be replaced by $f_1(d, r) = d$ and $f_2(d, r) = r$ respectively to simplify the computation of $\texttt{2d\_pos}$.  

The above embedding function constructs node and edge features by selecting values from a subset of the dimensions of $x$ and adding them to a bias vector, where both the subset of dimensions and the value of the vector are chosen based on the unique combination of values $v_1$ and $v_2$.  This ensures that the features $e$ injects are locally distributed along the dimensions of the input.  Though the features injected by $e$ take the form of waves, their frequencies---their only variable characteristic---are drawn from a discrete distribution.  

\subsection{Symbolic Program Design}
\label{sec:symbolic-program-design}

The symbolic programs used in this work are of type $G_2 \to G_2$.  Upon evaluation, all programs are applied to the same initial graph.  The primitive operations of the initial Domain-Specific Library (DSL) are such that symbolic programs may only make decisions based on the information contained in the graph they operate on.  Since the initial graph is uninformative, symbolic programs therefore only have access to information about their own semantics as it manifests in the graph elements they construct.  More detailed information on the operations of the initial DSL is available in the documentation of the \href{https://github.com/shlomenu/antireduce-graphs}{\texttt{antireduce-graphs}} library.

During program search, symbolic programs are considered novel if they generate a relational structure that is unique under isomorphism when symbolic features are not considered.  Isomorphism testing is performed with the VF2++ algorithm \citep{vf2++}.

\end{document}